\documentclass[letterpaper, 10 pt, conference]{ieeeconf}  

\IEEEoverridecommandlockouts                              
\usepackage{graphicx}
\usepackage{color,soul}
\usepackage{hyperref}
\usepackage{mwe}
\usepackage[caption=false,font=footnotesize]{subfig}
\usepackage{amsmath}
\usepackage{tabularx} 
\usepackage{multirow}

\newcolumntype{L}{>{\centering\arraybackslash}m{3cm}}
\usepackage{float}
\usepackage[thinc]{esdiff}
\usepackage{graphics} 
\usepackage[table]{xcolor}
\usepackage{algorithmic}
\usepackage[ruled,norelsize]{algorithm2e} 
\usepackage{caption}
\usepackage{graphics} 
\usepackage{epsfig} 
\usepackage{mathptmx} 
\usepackage{times} 
\usepackage{amsmath} 
\usepackage{amssymb}  
\usepackage[noadjust]{cite}
\usepackage{booktabs}
\usepackage{multirow}
\usepackage{algorithmic}
\usepackage[ruled,norelsize]{algorithm2e} 
\usepackage[noadjust]{cite}
\usepackage{censor}

\usepackage{microtype}
\renewcommand{\baselinestretch}{0.945}

\title{\LARGE \bf
Distributed Acoustic Localization Array\\ Deployed Using a Soft Everting Vine Robot
}

\author{Sebastian Lorca Godoy$^{1,2}$, Ciera McFarland$^{1}$, Michael Val$^{3}$, Antonio Alvarez Valdivia$^{3}$, \\Nathaniel Hanson$^{3}$, and Margaret McGuinness$^{1}$
\thanks{$^{1}$Department of Aerospace and Mechanical Engineering, University of Notre Dame, Notre Dame, IN 46556, USA. {\tt\footnotesize\{slorcago, cmcfarl2, mmcguinness\}@nd.edu}}%
\thanks{$^{2}$Pontifical Catholic University of Chile, 8320165 Santiago, Región Metropolitana, Chile. {\tt\footnotesize sdlorca@uc.cl}}
\thanks{$^{3}$\protect\raggedright Lincoln Laboratory, Massachusetts Institute of Technology, Lexington, MA 02421, USA. {\tt\footnotesize\{antonio.alvarezvaldivia, michael.val, nathaniel.hanson\}@ll.mit.edu}}
\thanks{DISTRIBUTION STATEMENT A. Approved for public release. Distribution is unlimited.
This material is based upon work supported by the Department of the Air Force under Air Force Contract No. FA8702-15-D-0001 or FA8702-25-D-B002. Any opinions, findings, conclusions or recommendations expressed in this material are those of the author(s) and do not necessarily reflect the views of the Department of the Air Force.
© 2026 Massachusetts Institute of Technology.
Subject to FAR52.227-11 Patent Rights - Ownership by the contractor (May 2014).
Delivered to the U.S. Government with Unlimited Rights, as defined in DFARS Part 252.227-7013 or 7014 (Feb 2014). Notwithstanding any copyright notice, U.S. Government rights in this work are defined by DFARS 252.227-7013 or DFARS 252.227-7014 as detailed above. Use of this work other than as specifically authorized by the U.S. Government may violate any copyrights that exist in this work.}
}

\begin{document}

\maketitle
\thispagestyle{empty}
\pagestyle{empty}

\begin{abstract}
Soft robot exteroception is increasingly being explored for a variety of field applications. In this work, we present a sound-based system for localizing disaster victims in confined and unstructured environments, based on a distributed acoustic sensing architecture embedded along the body of a soft everting vine robot. We propose a dynamic Steered Response Power with Phase Transform framework that supports both far-field direction-of-arrival estimation and near-field three-dimensional source localization as the robot approaches the sound source. To better understand the design and control space related to localizing sound using a soft, shape-morphing robot body, we conduct experiments measuring the accuracy of these methods for a five-microphone array attached to the robot body using three placements relative to the outer membrane of the robot (inside the pressurized body, inside the inner tail, and outside the outer wall) and in four robot configurations (linear, double linear, circular, and sinusoidal). We measure the change in accuracy as the signal-to-noise ratio, the direction of approach, and the distance of the sound source from the center of the array change. Finally, we demonstrate a vine robot growing into an arbitrary shape while carrying microphones along its outer wall, and show that a sound source located with the array's near field can be localized with high accuracy after only three microphones have everted from the robot body. These results highlight the potential of distributed acoustic sensing for reliable victim localization using soft growing robots.

\end{abstract}


\section{Introduction}
\label{sec:intro}

Soft robots have advantages over traditional rigid robots in their adaptability to the environment. However, their inherent flexibility poses unique challenges when using them as a sensory platform \cite{Lee2017SoftRobotReview}. Rigid sensors can inhibit the adaptable motions that make soft robots desirable, which has led to a growing interest in integrating fully soft sensors into these robots \cite{qu2024sensing}. However, soft sensors carry many of the same modeling challenges as soft robots and can require additional compensation to account for drift \cite{thuruthel2019shapesensing}. Alternatively, soft robots can also employ rigid sensors if they are low-profile, leading many smart soft robotic systems to still contain some amount of rigid components~\cite{hegde2023sensing}.

\begin{figure}[!t]
    \centering
    \includegraphics[width=\linewidth, page=1]{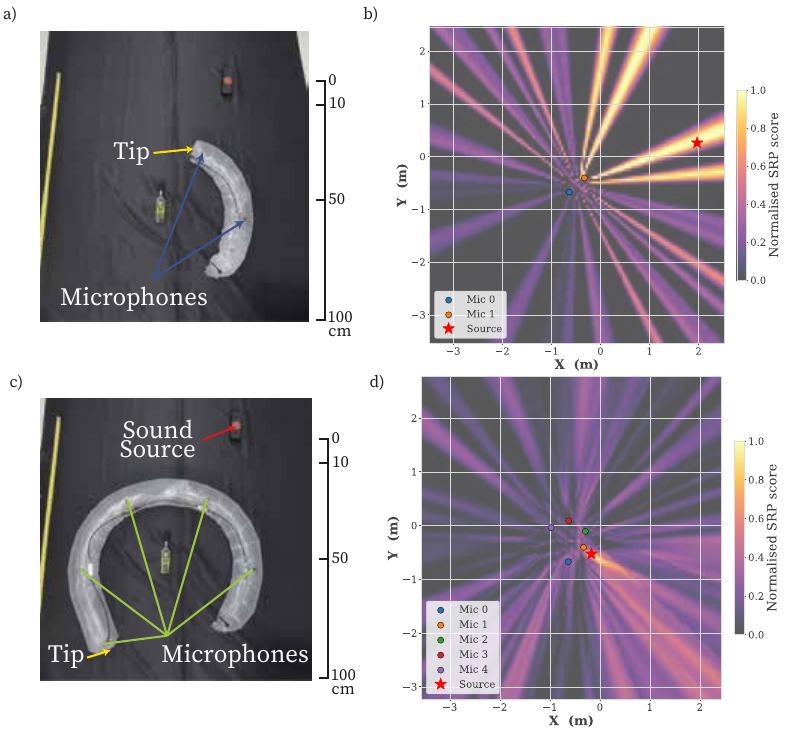}
    \caption{A vine robot with (a) two and (c) five microphones progressively everted along its body, and the corresponding SRP-PHAT spatial power maps for (b) two microphones, showing broad and ambiguous power distribution across the candidate grid, and (d)  five microphones, showing a concentrated power peak in the vicinity of the true source location, demonstrating the accuracy improvement in near-field as the array aperture grows during eversion.}
    \label{fig:usar_teaser}
    \vspace{-2.0em}
\end{figure}

Urban search and rescue (USAR) operations are a prime example of an environment where soft robots have unique potential. These environments are cluttered, dangerous, and often have no known map \cite{murphy2017disaster, frost2025rubblesim}, so soft robots that flexibly navigate the space could outperform traditional, rigid robots. The few soft robots that have been demonstrated in relevant environments -- active scope cameras \cite{Fukuda2014verticalASC,Fujikawa2019asc} and everting vine robots \cite{der2021roboa,mcfarland2024field} -- only carried sensors at the tip or forwardmost point of the robot body. These prior works focused on challenges with robot design, locomotion, and teleoperation. However, in order to be useful as a search and rescue tool, a robot may need to carry a wide array of sensors, such as microphones, cameras, and inertial measurement units. 

The unique geometry of everting vine robots, which move through their environment by turning their bodies inside-out and ``growing'' from the tip~\cite{al2025tip}, enables them to be used as continuous sensor arrays with reconfigurable geometries~\cite{gruebele2021distributed, bryant2022tactile, mitchell2023soft, laudenslager2026evaluating}. Given the correct modeling of signals perceived at multiple points along the robot's body, vine robots could be used to pinpoint environmental phenomena, such as a trapped person calling out for help.

 Existing disaster victim localization systems often use a series of microphones wired to each other and placed around or near the surface of a rubble pile to detect human speech or tapping noises \cite{murphy2017disaster}. In this work (Figure~\ref{fig:usar_teaser}), we sensorize a vine robot with microphones along its length which enable it to deploy and reconfigure the microphone array for improved sound source localization (SSL). To our knowledge, no work has investigated steerable microphone array deployment and reshaping via a robot. Therefore, there are many unknowns regarding how environmental noise, microphone placement, robot configuration, and sound source location will influence the accuracy of a vine robot's localization algorithm.  

The main contributions of this work are as follows:
\begin{itemize}
    \item A distributed acoustic sensing array embedded along the body of an everting vine robot for sound-based victim localization in USAR environments.
    
    \item Experimental characterization of SSL accuracy across vine-robot-generated array geometries under near-field and far-field source conditions, quantifying the effects of environmental noise, microphone placement, robot configuration, and sound source location.

    \item Demonstration of an everting vine robot growing while localizing a sound source with increasing accuracy as more microphones deploy along the robot's body.
    
\end{itemize}

\section{Related Work}
\label{sec:related_work}

Robot audition refers to the study of machine hearing in the context of robotics, where acoustic perception accounts for challenges that are largely absent in controlled, static environments: self-generated noise from motors and actuators, moving or deforming microphone arrays, real-time computational constraints, and the need to integrate auditory information with other sensors~\cite{okuno2015audition}. A variety of works have explored sound source localization using multi-microphone arrays \cite{yang2024kidnappable, sasaki2018online, valin2003robust}. Binaural approaches, which emulate the two-ear hearing of biological organisms, have been explored for robot audition while using a minimal number of microphones~\cite{keyrouz2006binaural}.

These robotic systems seek to achieve sound source localization (SSL) -- the process of estimating the direction of arrival (DOA) or the three-dimensional position of an acoustic source from the signals recorded by a microphone array. The literature on SSL can be broadly divided into time-delay estimation methods, spectral subspace methods, and, more recently, deep-learning  approaches~\cite{adavanne2018}.

The generalized cross-correlation with phase transform (GCC-PHAT) algorithm forms the backbone of many practical SSL systems~\cite{knapp1976gcc}. GCC-PHAT estimates the time difference of arrival (TDOA) between each microphone pair by applying a phase transform that whitens the cross-power spectrum, i.e., normalizes its magnitude so that all frequency components contribute equally, before computing the inverse Fourier transform. The whitening step reduces the influence of spectral coloration and reverberation, making GCC-PHAT a robust building block for modern localization. When multiple microphone pairs are available, individual TDOA estimates can be used to recover the source position.

Steered response power (SRP)~\cite{dibiase2000srp} methods treat localization as a grid search over a candidate position space: for each candidate, the array response is steered toward that position by applying the time delays expected for a source located at that point, effectively aligning the microphone signals. The resulting output power is then accumulated across microphone pairs and frequencies. The steered response power with phase transform (SRP-PHAT) variant~\cite{dibiase2000srp} applies the phase transform (PHAT) weighting in the frequency domain, normalizing the magnitude of the cross-power spectrum so that only phase information is retained. This yields robustness to reverberation comparable to GCC-PHAT while producing an explicit spatial map whose global maximum indicates the source location. The principal advantage of SRP-PHAT is its ability to operate with arbitrary, uncalibrated array geometries, a property that is critical when microphones are deployed along the everting body of a vine robot. The main disadvantage is computational complexity: in-depth search scales as $O(N_\text{cand})$, where $N_\text{cand}$ is the number of grid points in the map, though efficient fast Fourier transform-based implementations and branch-and-bound pruning have partially mitigated this~\cite{dibiase2000srp}.

The multiple signal classification (MUSIC) algorithm uses the eigenstructure of the array covariance matrix to achieve super-resolution DOA estimation~\cite{schmidt1982ssl_music}. The covariance matrix of the received signals is decomposed into two orthogonal subspaces: the signal subspace, spanned by the steering vectors of the active sources, and the noise subspace, spanned by the remaining eigenvectors. MUSIC exploits the fact that the steering vector of a true source is orthogonal to the noise subspace, and constructs a pseudo-spectrum, i.e., a function of candidate direction whose value is inversely proportional to the projection of the candidate steering vector onto the noise subspace. The pseudo-spectrum produces sharp peaks at the true source directions, resulting in higher angular resolution than beamforming-based methods such as SRP-PHAT.

This approach performs well in low-noise, anechoic conditions with a well-calibrated array, but its performance degrades significantly when array manifold errors (inconsistencies between the assumed and true steering vectors arising from geometric uncertainty) are present, which is a concern for flexible or partially occluded arrays, or when the number of sources is unknown. Furthermore, MUSIC requires that the number of sources be specified \textit{a priori}, which is often impractical in unstructured deployment environments.

The estimation of signal parameters via the rotational invariance technique (ESPRIT)~\cite{roy1989ssl_esprit} avoids the computationally expensive spatial search of MUSIC by using a shift invariance structure between two identical sub-arrays. By solving a generalized eigenvalue problem, ESPRIT yields closed-form DOA estimates at a lower computational cost than grid-based methods. However, ESPRIT imposes a strict requirement on array geometry: the array must consist of pairs of sensors with a constant displacement vector (\emph{doublets}). This structural constraint makes ESPRIT poorly suited to the vine robot setting, where the microphone layout along the robot body is determined by the path taken during growth and cannot, in general, be guaranteed to satisfy the doublet condition.

Given the irregular and environment-dependent geometry of a vine robot's microphone array, we select SRP-PHAT as the most natural starting point for our work, which focuses on the fusion of these algorithmic techniques and a deformable, deployable microphone array for SSL.
\section{Sound Source Localization Algorithm}
\label{sec:ssl_algorithm}

In the SRP-PHAT algorithm~\cite{dibiase2000srp},
for each microphone pair $(m_k, m_l)$ drawn from an array of $M$ microphones, the GCC-PHAT~\cite{knapp1976gcc} is computed as

\begin{equation}
    R_{m_k m_l}(\tau) = \int_{-\infty}^{\infty}
        \frac{M_k(\omega)\,M_l^*(\omega)\,e^{j\omega\tau}}
             {|M_k(\omega)\,M_l(\omega)|}
    \,d\omega,
\end{equation}

\noindent where $\omega = 2\pi f$ is the angular frequency in radians per second, $j$ is the imaginary number, $M_k(\omega)$ and $M_l(\omega)$ are the Fourier transforms of the signals recorded at microphones $m_k$ and $m_l$, respectively, $(\cdot)^*$ denotes complex conjugation, and $\tau$ is the time delay over which the correlation is evaluated.

For a candidate source position $\mathbf{x}$ in the set $\mathcal{G}$ of 
candidate positions, the theoretical TDOA between microphones $m_k$ and $m_l$ is

\begin{equation}
    \tau_{kl}(\mathbf{x}) =
        \frac{\|\mathbf{x} - \mathbf{x}_k\| - \|\mathbf{x} - \mathbf{x}_l\|}{c},
\end{equation}

\noindent where $\mathbf{x}_k$ and $\mathbf{x}_l$ are the known positions of the two microphones, $c$ is the speed of sound, and the subscript $kl$ denotes the microphone pair $(m_k, m_l)$. The SRP-PHAT functional ($P'_n$) accumulates the GCC-PHAT evaluated at the predicted delay across all $\binom{M}{2}$ unique microphone pairs,

\begin{equation}
    P'_n(\mathbf{x}) =
        \sum_{k=1}^{M} \sum_{l=k+1}^{M}
        R_{m_k m_l}\!\left(\tau_{kl}(\mathbf{x})\right),
\end{equation}

\noindent where $n$ indexes the time frame over which the signals are processed, producing a spatial power map whose peaks indicate likely source locations. The estimated source position is then obtained as

\begin{equation}
    \mathbf{x}_s = \arg\max_{\mathbf{x} \in \mathcal{G}}\, P'_n(\mathbf{x}).
\end{equation}

The candidate set $\mathcal{G}$ is usually defined as a uniform spatial grid. For planar localization, $|\mathcal{G}| = |\mathcal{G}^{(1)}|\times |\mathcal{G}^{(2)}|$, where $|\mathcal{G}^{(1)}|$ and $|\mathcal{G}^{(2)}|$ are the number of points along the length and width,  respectively~\cite{grinstein2024ssl}.

A key property of SRP-PHAT is that the same algorithm applies to both near-field and far-field conditions, with the operating regime determined entirely by the definition of the candidate set $\mathcal{G}$ and the corresponding delay model $\tau_{kl}(\mathbf{x})$. 

In the far-field case, the plane-wave approximation simplifies the delay model: since the source is effectively at infinity, $\tau_{kl}$ depends only on the DOA and not on the source range. The candidate set $\mathcal{G}$ is therefore defined as a set of unit direction vectors, parameterized by a single azimuth angle $\theta$ in the two-dimensional case, or by azimuth and elevation $(\theta$, $\phi$) in three dimensions, and $\tau_{kl}$ reduces to

\begin{equation}
    \tau_{kl}(\theta) = \frac{(\mathbf{x}_k - \mathbf{x}_l)\cdot\hat{\mathbf{u}}(\theta)}{c}
\end{equation}

\noindent where $\hat{\mathbf{u}}(\theta)$ is the unit vector pointing toward the candidate direction $\theta$~\cite{grinstein2024ssl}. In this region, SRP-PHAT produces an angular power map over $\mathcal{G}$, and $\mathbf{x}_s$ represents an estimated direction rather than a position. A circular grid with uniformly spaced angles is a common choice for $\mathcal{G}$ in the two-dimensional far-field case~\cite{grinstein2024ssl}.


\section{Experiments and Results}
\label{sec:experiments}

In this section, we present the setup and results of three sets of experiments examining how microphone placement, robot configuration, and sound source distance impact the accuracy of the robot's SSL capabilities as the signal-to-noise ratio (SNR) and DOA change.

\subsection{Experimental Setup}
 
For these experiments, we built a $20.3$~cm lay-flat diameter vine robot from low-density polyethylene (LDPE) with nominal thickness $0.1$~mm (Hudson Exchange, Ohio, USA). The electronics array consisted of five ICS43434 microphones (TDK InvenSense, California, USA), controlled by a Teensy 4.1 microcontroller unit (PJRC, Oregon, USA). For all configurations, we spaced consecutive microphones $30$~cm apart as measured along the arc length of the robot body.

\begin{figure}[tb]
    \centering
    \vspace{0.5em}
    \includegraphics[width=\linewidth]{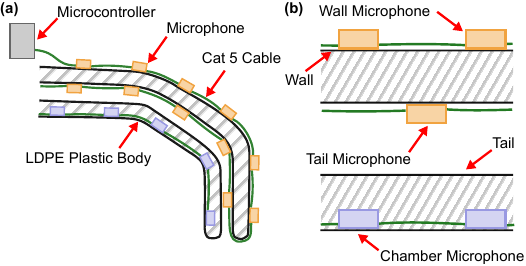}
    \caption{Microphone placement diagram. (a) Schematic showing cross-sectional view of the vine robot body with microphones distributed along its length, connected by a Cat 5 cable to the microcontroller at the robot's base. (b) Zoomed-in view illustrating the mounting configurations of microphones embedded in the pressurized body, the tail, and the wall.}
    \label{fig:test_setup}
    \vspace{-1.0em}
\end{figure}

We evaluated three microphone placements relative to the outer membrane of the vine body, shown in Figure~\ref{fig:test_setup}, and carried out all placement experiments sequentially using the same vine body. For the wall placement, we attached the microphones externally to the surface of the fully everted vine. For the tail placement, we first mounted the microphones on the wall at the end of the fully everted vine, then manually inverted the robot until the tip microphone reached the base of the body, placing all the microphones inside the non-everted tail region. For the pressurized body placement, we inserted the microphones inside the inflated body and secured them with adhesive prior to heat-sealing the vine. After data collection, we removed the seal, unmounted the microphones, and resealed the body for the following experiments.

\begin{figure}[tb]
    \centering
    \includegraphics[width=\linewidth]{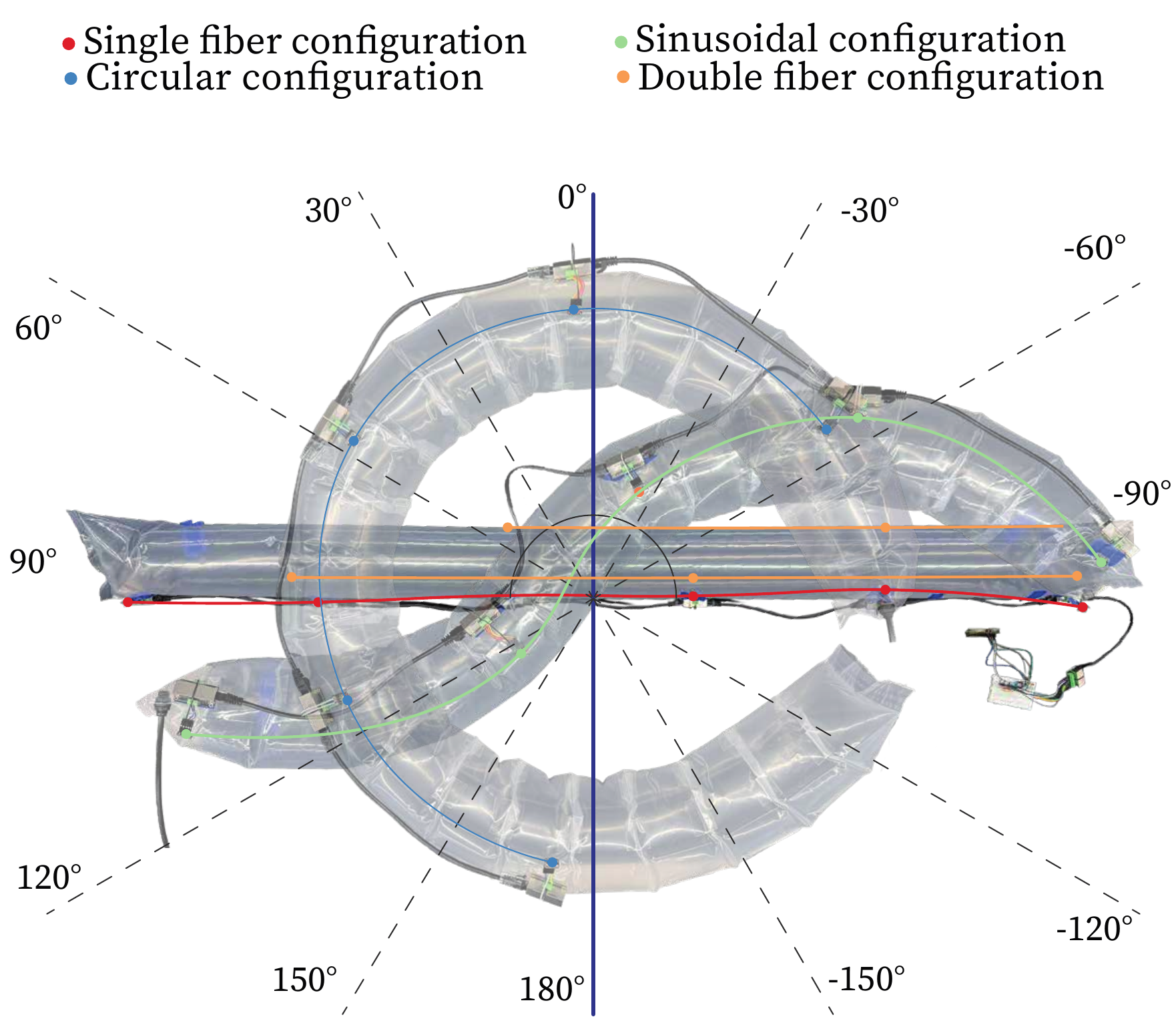}
    \caption{Evaluated array configurations (2D) with five microphones mounted per geometry and reference angles used for experiments.}
    \label{fig:setup shapes}
    \vspace{-2.0em}
\end{figure}

We evaluated four robot configurations, as shown in Figure~\ref{fig:setup shapes}. In the single fiber configuration, all five microphones were arranged collinearly along the robot body, spaced $30$~cm apart along the arc length. In the double fiber configuration, five microphones were distributed across two parallel fibers in an interleaved arrangement, with three on one fiber and two on the other, resulting in a lateral separation of $14$~cm between fibers. In the circular configuration, the vine body was formed into a closed loop with a radius of $33$~cm, distributing the microphones uniformly around the circumference. In the sinusoidal path, using a contraction ratio of $80\%$, for every $10$~cm of body length along the primary axis, the opposing fold reached $8$~cm, producing the curved shape.

We used a SoundCore~2 speaker (Anker, Shenzhen, China) playing a standard Gaussian white noise signal to act as the acoustic source across all experiments. We collected audio data at a sampling rate of $44.1$~kHz and processed it in frames of $2048$ samples, corresponding to a window duration of approximately $46.6$~ms, which provides a frequency resolution of $\Delta f = 44100/2048 \approx 21.5$~Hz per bin. Additionally, we conducted the experiments in a standard laboratory room with a non-negligible ambient noise floor, originating from ventilation ducts and electrical equipment operating continuously during the experiments. We measured the sound source level and the ambient noise level independently at the center of the array using a calibrated sound level meter (SL720, Tadeto), and computed the SNR as

\begin{equation}
    \text{SNR} = 10 \log_{10} \left( \frac{P_{\text{total}} - N}{N} \right)
\end{equation}

\noindent where $P_{\text{total}}$ and $N$ are the measurements of the whole system and the noise levels, in A-weighted readings. Since both measurements were obtained with the same A-weighted meter, the SNR is reported in dB and represents the ratio of source to noise pressure levels at the array center. We varied the SNR across experimental conditions by increasing the speaker output power while keeping the environment and its noise sources unchanged, maintaining a constant $45$~db(A) throughout. For each experimental condition discussed in the following subsections, we collected a total of $300$ adjacent frames and calculated the mean absolute angular or position error, along with its standard deviation, across all frames to produce a representative estimate of localization accuracy.

We determined the three-dimensional position of each microphone and the sound source using PhaseSpace (PhaseSpace Inc., San Leandro, USA), an optical motion capture system that tracks the position of active LED markers via multiple calibrated cameras. We attached LED markers to each microphone and to the speaker enclosure prior to each trial, allowing us to recover the exact spatial configuration and the real source position for each experimental condition. All experiments were conducted in two dimensions: the candidate position space was restricted to the $xy$-plane, the sound source was placed at the same height as the microphone array, and all microphones were positioned at the same height relative to each other, so that out-of-plane contributions to the TDOA were minimal. For all near-field experiments, with the exception of Section~\ref{subsec:distance}, we defined the candidate position grid as a $3\times 3$~m space centered on the center of the array, with a uniform resolution of $5$~cm, resulting in $3600$ candidate positions evaluated per frame.

\subsection{Microphone Placement Experiments}
\subsubsection{Far-Field}
\label{subsec:ff_placement}

To evaluate the effect of microphone placement on far-field source azimuth estimation accuracy under various SNRs, we placed the speaker at a fixed azimuth of $290^\circ$ for the circular configuration at a distance of $2$~m from the center of the array. We tested the three microphone placements shown in Figure~\ref{fig:test_setup}.

\begin{figure}[tb]
    \centering
    \includegraphics[page=1, width=0.9\linewidth]{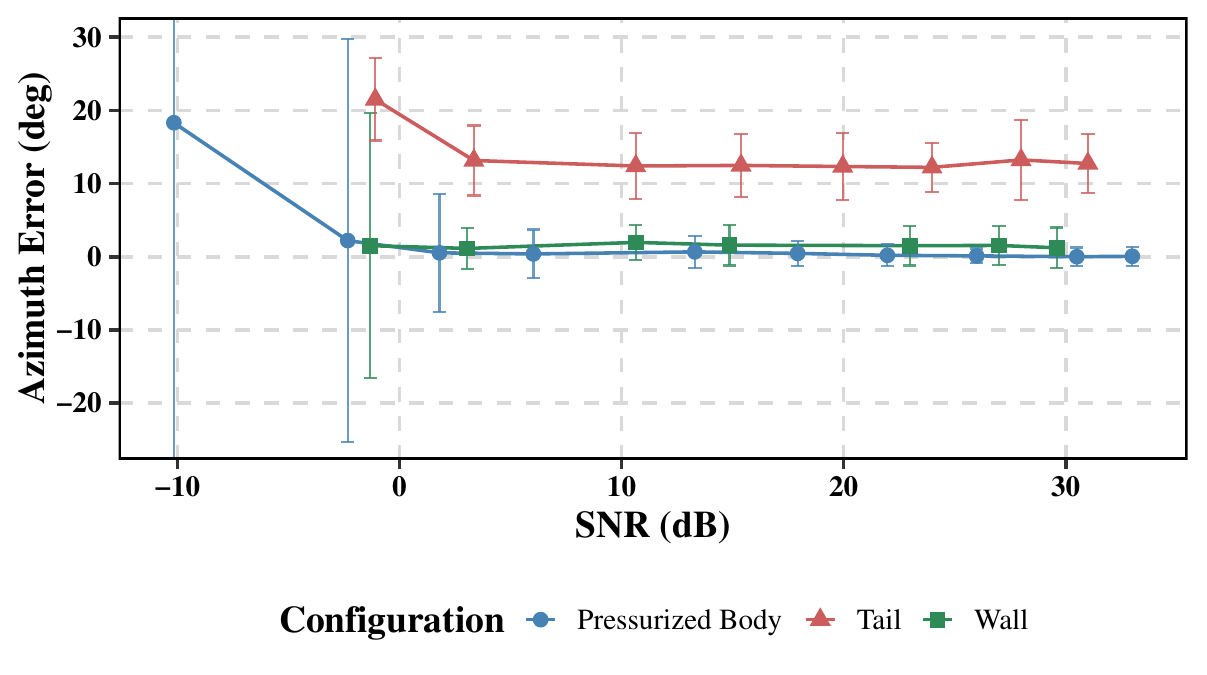}
    \caption{Mean absolute azimuth localization error as a function of signal-to-noise ratio (SNR) for a circular robot configuration for far-field measurements. Results are shown for microphones placed inside the pressurized body, inside the tail, and outside the wall. Across most SNR levels, we observe the lowest error when the microphones are inside the pressurized body, and the highest error when the microphones are placed inside the tail.}
    \label{fig:exp_acc_ff}
    \vspace{-2.0em}
\end{figure}

Figure~\ref{fig:exp_acc_ff} reports the resulting localization error curves.
There is a clear performance hierarchy among the three mounting configurations. Microphones placed inside the tail fail to produce accurate DOA estimates, exhibiting a near-constant mean absolute azimuth error of approximately $10.4^\circ$ and a standard deviation of $4.6^\circ$, considering every measurement taken. We attribute this to the acoustic path geometry at the tip: sound waves reach the tail microphones through two competing paths: transmission through the double LDPE wall and direct coupling through the open gap at the tip of the robot body, producing a superposition of attenuated and delayed wavefronts that distorts the SRP-PHAT algorithm, regardless of the source level.

Wall-mounted and pressurized body microphones both converge to low mean absolute errors as SNR increases above $0$~dB, where the source signal is at least as strong as the ambient noise floor, averaging $1.49^\circ$ and $0.51^\circ$ azimuth error, respectively. Below this threshold, localization becomes increasingly unreliable for both configurations, increasing in both mean and standard deviation of error as SNR decreases. 
Overall, microphones placed within the pressurized body provide the most consistent and accurate localization performance, followed by wall-mounted microphones, with tail-mounted microphones performing substantially worse across the entire SNR range. We attribute the performance gap between the pressurized body and wall configurations to the acoustic non-uniformity introduced by the vine body itself: for wall-mounted microphones, the inflated LDPE tube partially obstructs the acoustic path in a way that varies with the position of each individual microphone along the body, introducing configuration-dependent amplitude and phase distortions. For pressurized body microphones, the LDPE wall acts as a spatially uniform acoustic barrier that attenuates the incoming signal consistently across all microphones, preserving the inter-microphone phase differences on which SRP-PHAT relies.

\subsubsection{Near-Field}
\label{subsec:nf_placement}

To evaluate the effect of microphone placement on near-field source position estimation accuracy for various SNRs, we placed the speaker at a fixed azimuth of $290^\circ$ and a distance of $0.75$~m from the center of the array. Otherwise, the experimental conditions were identical to those in Section~\ref{subsec:ff_placement}.

\begin{figure}[tb]
    \centering
    \includegraphics[width=0.9\linewidth]{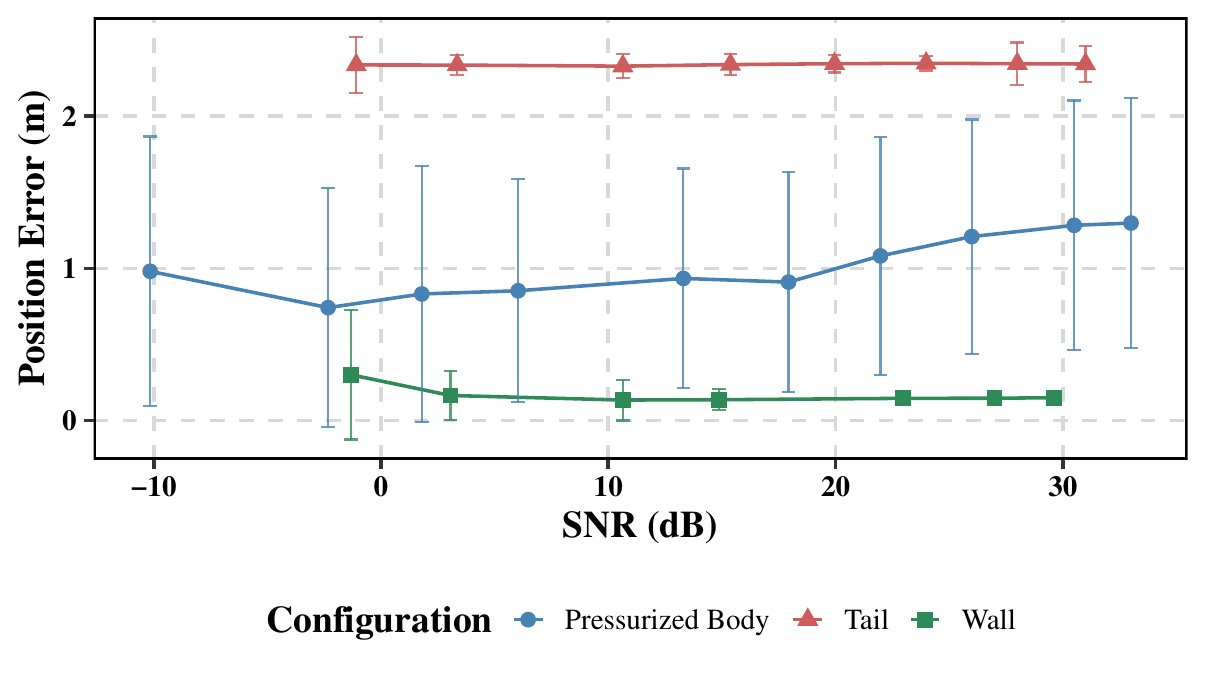}
    \caption{Mean absolute Euclidean distance localization error as a function of SNR for a circular robot configuration for near-field measurements. Results are shown for the same microphone placements as in Figure~\ref{fig:exp_acc_ff}. Across all SNR levels, we observe the lowest error when the microphones are placed outside the wall, and the highest error when the microphones are placed inside the tail.}
    \label{fig:exp_acc_nf}
    \vspace{-1.9em}
\end{figure}

Figure~\ref{fig:exp_acc_nf} reveals that the tail configuration again fails to produce reliable position estimates, with errors saturating at $2.3$~m across all SNR levels, corresponding to the maximum distance from the source to the far boundary of the $3 \times 3$~m candidate grid. This indicates the tail microphones are unable to resolve sound source position under near-field conditions, likely for the same acoustic path reasons discussed previously.

The results also show a reversal in the relative performance of the pressurized body and wall-mounted configurations compared to the far-field case.  Wall-mounted microphones achieve the highest localization accuracy, with a mean absolute position error of $0.17$~m and a standard deviation of $0.13$~m averaged across all SNR levels. Pressurized body microphones, which outperform all placements in the far-field experiment, exhibit substantially degraded accuracy in the near-field, with a mean absolute position error of $1.012$~m and a standard deviation of $0.788$~m averaged across all SNR levels. Wall-mounted microphones therefore achieved a mean absolute error $83.3\%$ lower than pressurized body microphones
, with an $84.0\%$ reduction in standard deviation
, indicating a substantially more consistent estimator under near-field conditions.

In the near-field, SRP-PHAT relies on wavefront curvature to jointly estimate source range and DOA, which requires that the phase differences between microphones accurately reflect the spherical wavefront geometry. Pressurized body microphones receive a spatially uniform but attenuated version of the incident field, which preserves inter-microphone phase relationships at far-field distances where the wavefront is approximately planar. At near-field distances, however, the curvature of the wavefront produces subtler and more spatially varying phase differences that are more susceptible to the attenuation introduced by the LDPE wall. Wall-mounted microphones, being in more direct acoustic contact with the propagating field, better preserve the curvature information required for accurate near-field range estimation.

\subsection{Robot Configuration Experiments}
\subsubsection{Far-Field}
\label{subsec:ff_shapes}

\begin{figure}[tb]
    \centering
    \includegraphics[width=0.9\columnwidth]{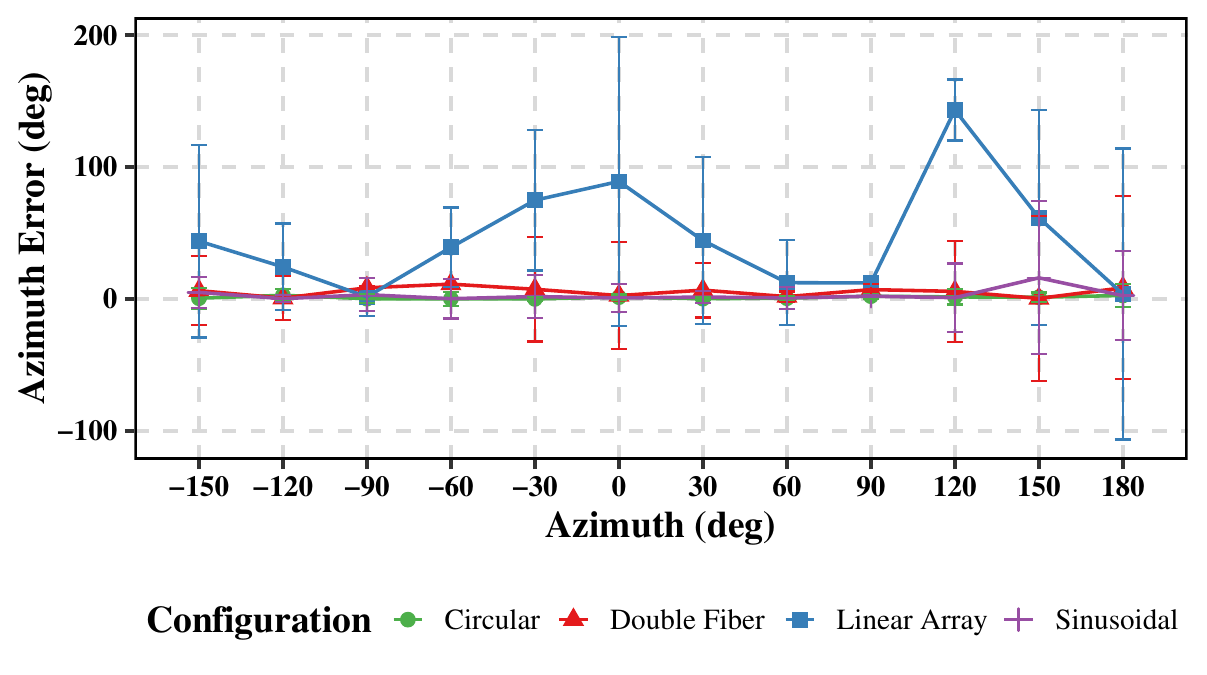}    
    \caption{Mean absolute azimuth localization error as a function of azimuth for the linear, double linear, circular, and sinusoidal configurations for far-field measurements. The linear array has the largest error at most azimuths. The other three geometries have significantly lower error across the azimuth range, with the circular configuration having the lowest overall.}
    \label{fig:shape_ff}
    \vspace{-1.5em}
\end{figure}

To evaluate the effect of robot configuration on far-field source azimuth estimation accuracy across the full range, we positioned the speaker at a fixed distance of $2$~m from the center of the array and swept across target azimuths from $-180^\circ$ to $180^\circ$ in increments of $30^\circ$, with a constant SNR of $13$~dB. We tested the four robot configurations shown in Figure~\ref{fig:setup shapes}. Figure~\ref{fig:shape_ff} reports the resulting error profiles as a function of target azimuth, revealing the angular regions where each configuration performs best and where localization ambiguities are most pronounced.

 The single fiber configuration produced the largest errors, with a mean absolute azimuth error of $45.9^\circ$ and a standard deviation of $52.2^\circ$ across all tested azimuths, indicating both poor accuracy and highly inconsistent estimates. This is consistent with the known limitations of collinear arrays: when all microphones lie along a single axis, the TDOA between any pair depends only on the projection of the source direction onto that axis, making the SRP-PHAT map symmetric about the array axis and unable to distinguish sources on opposite sides. This front-back ambiguity manifests as elevated errors near the azimuths aligned with the array axis, with larger inaccuracy the farther the azimuth is from $90^\circ$ and $-90^\circ$. The high standard deviation reflects the erratic nature of these estimates, where the algorithm alternates between the correct solution and its mirror image depending on noise realizations.

The double fiber configuration reduced the mean error to $5.6^\circ$, an $87.9\%$ improvement over the single fiber baseline, 
though its standard deviation remained large at $27.0^\circ$. This suggests that while the additional fiber breaks the front-back ambiguity on average, estimates remain susceptible to large deviations at specific azimuths. The sinusoidal configuration further reduced the mean error to $2.9^\circ$ with a standard deviation of $16.7^\circ$, a $93.6\%$ reduction relative to the single fiber case, reflecting the benefit of the non-collinear microphone distribution introduced by the sinusoidal geometry. Both the double fiber and sinusoidal configurations exhibited localised error increases at azimuths aligned with their linear or quasi-linear segments, where collinear microphone pairs contribute ambiguous TDOA estimates that partially corrupt the accumulated SRP-PHAT functional.

The circular configuration achieved the lowest mean error of $1.1^\circ$ and the lowest standard deviation of $3.9^\circ$, representing a $97.6\%$ decrease in mean error relative to the single fiber configuration and a $62.1\%$ decrease relative to the sinusoidal configuration. The substantially lower standard deviation of the circular configuration, $4.3\times$ smaller than that of the sinusoidal arrangement ($3.9^\circ$ vs. $16.7^\circ$), indicates that the circular array produces not only more accurate but more consistent estimates across the full azimuthal range. 

\subsubsection{Near-Field}

To evaluate the effect of robot configuration on near-field source position estimation accuracy across the full azimuthal range, we positioned the speaker at a fixed distance of $0.5$~m from the center of the array and swept across target azimuths from $-180^\circ$ to $180^\circ$ in increments of $30^\circ$, with a constant SNR of $13$~dB.
We excluded the linear configuration from the near-field evaluation, as limited aperture produces insufficient wavefront curvature discrimination. Otherwise, the experimental conditions were identical to those in Section~\ref{subsec:ff_shapes}.

\begin{figure}[tb]
    \centering
    \includegraphics[width=0.9\linewidth]{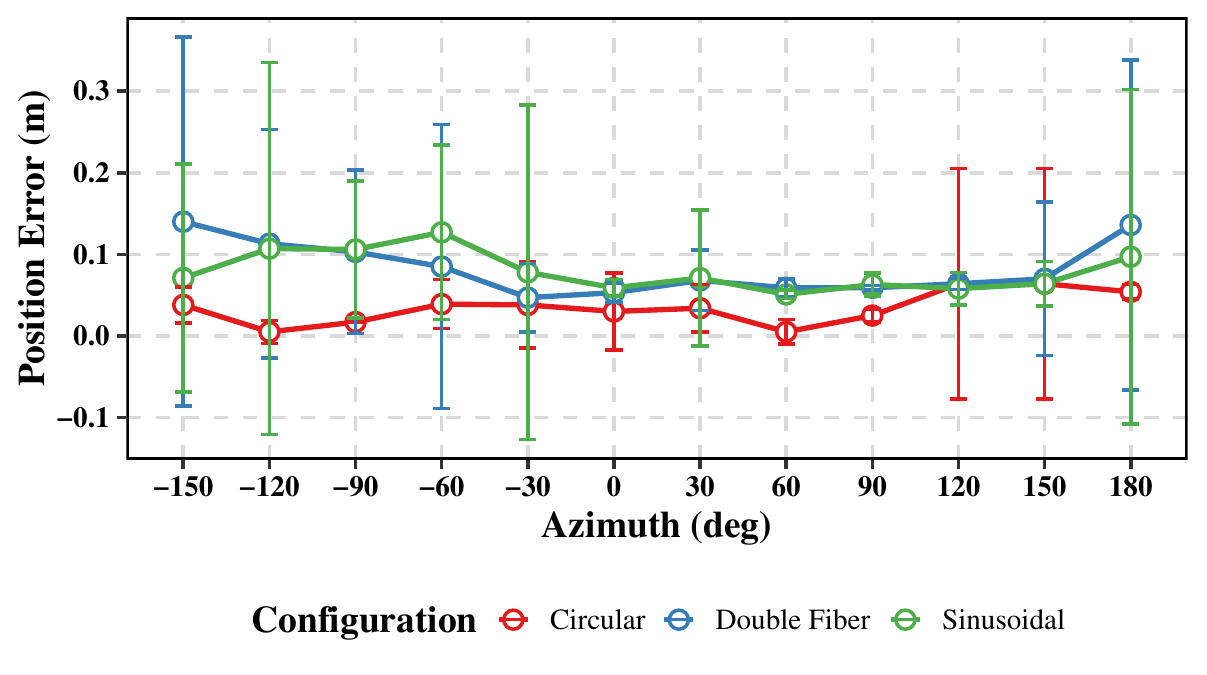}
    \caption{Mean absolute Euclidean distance localization error as a function of azimuth angle for near-field measurements. Results are shown for the double linear, circular, and sinusoidal configurations. Here, we again observe that the circular array has the lowest error across the azimuth range.}
    \label{fig:shape_nf}
    \vspace{-1.5em}
\end{figure}

Figure~\ref{fig:shape_nf} reports the resulting error profiles as a function of target azimuth. Across all angles, the circular array achieved the lowest mean error of $0.034$~m with a standard deviation of $0.042$~m, outperforming both the double fiber ($0.083$~m, $\sigma = 0.087$~m) and sinusoidal ($0.079$~m, $\sigma = 0.094$~m) configurations by $58.8\%$ and $56.6\%$ respectively. The double fiber and sinusoidal configurations performed comparably in mean error, differing by less than $5\%$, though the sinusoidal arrangement exhibited a marginally higher standard deviation, suggesting slightly less consistent estimates across trials.

The advantages of the circular configuration in the near-field are consistent with the far-field shape results of Section~\ref{subsec:ff_shapes}, where the circular array also achieved the lowest error, and supports the interpretation that greater aperture is the dominant factor governing accuracy regardless of the operating regime: far- and near-field.

\subsection{Sound Source Distance Experiment}
\label{subsec:distance}

\begin{figure}[tb]
    \centering
    \includegraphics[width=0.9\linewidth]{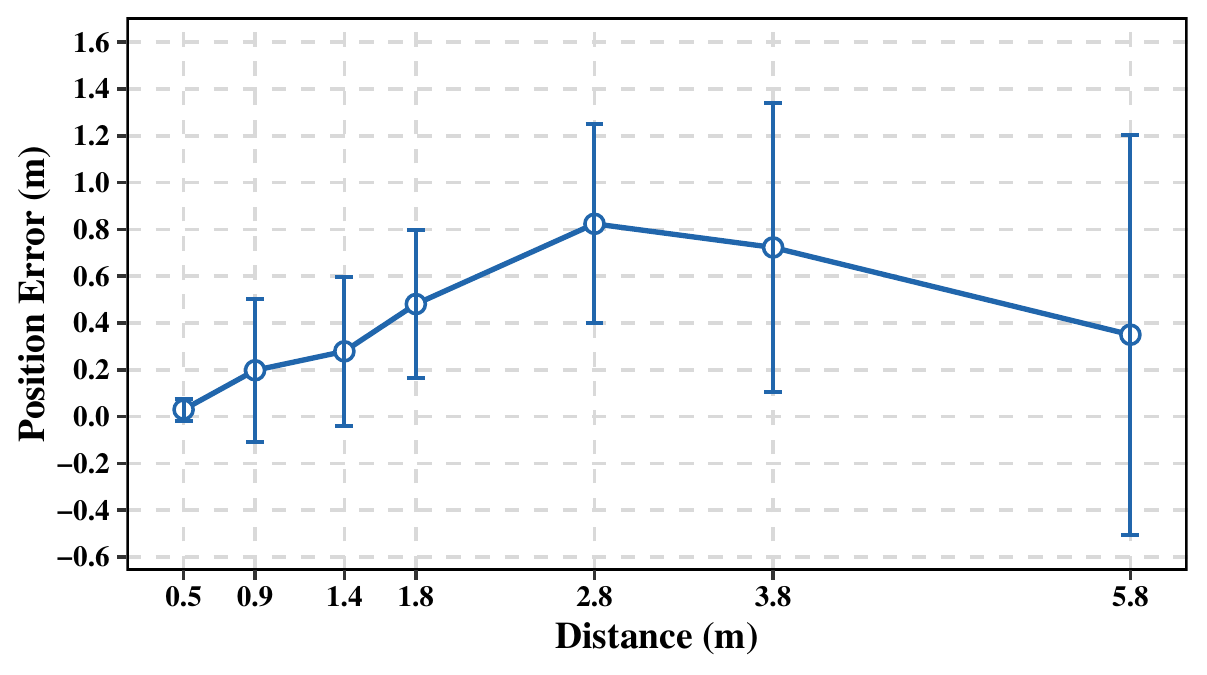}
    \caption{Mean absolute Euclidean distance localization error as a function of the distance between the source and the center of the array at $60^\circ$ in a circular configuration for near-field measurements. Error is lowest when the sound source is closest to the microphone. It increases with distance before eventually decreasing at the longest distances tested.}
    \label{fig:distance_nf}
    \vspace{-1.5em}
\end{figure}

To evaluate the effect of source distance on near-field source position estimation accuracy, we positioned a speaker at a fixed azimuth of $60^\circ$ and swept across distances of $0.5$~m, $0.9$~m, $1.4$~m, $1.8$~m, $2.8$~m, $3.8$~m, and $5.8$~m from the center of the array. These distances were determined by physical constraints of the laboratory space.
We selected the circular array because it provides the most spatially uniform aperture coverage among the configurations tested. Figure~\ref{fig:distance_nf} shows the resulting error profile as a function of source distance for a circular array of radius $r=0.33$~m, evaluated at a fixed azimuth of $60^\circ$. 

Within the near-field region, the mean position error increases monotonically with source distance: from $0.03$~m at $0.5$~m (RMSE~$= 0.056$~m) to $0.198$~m at $0.9$~m (RMSE~$=0.364$~m), reaching $0.279$~m at $1.4$~m (RMSE~$=0.423$~m). This trend reflects the progressive reduction in wavefront curvature as the source recedes, which reduces the clarity of the spatial power map and makes range estimation increasingly uncertain. The closest measurement at $0.5$~m achieves a mean error of $0.03$~m, or $6.0\%$ of the true source distance, demonstrating that SRP-PHAT with a circular vine robot array can localize near-field sources with high accuracy when the source is closer to a microphone than the largest inter-microphone distance.

Beyond the $1.5$~m boundary, the error profile becomes non-monotonic, with the mean error peaking near $2.8$~m before decreasing at larger distances. This reduction is not indicative of improved localization reliability; the standard deviation grows substantially across this range, and the RMSE remains above $0.92$~m for all measurements beyond $2.8$~m, indicating that individual estimates become increasingly dispersed as the source moves into the far-field. As the wavefront approaches planarity, the power map loses sensitivity in the range dimension and the peak position wanders across the candidate grid, producing a high-variance estimator whose mean error is not a meaningful measure of accuracy.

Taken together, these results suggest that the near-field operating range for this circular configuration is approximately $0.5$~m to $1$~m, within which the mean error remains below $0.4$~m and the RMSE grows predictably with distance.

\section{Demonstration}
\label{sec:demonstration}

Having characterized the effect of array shape and microphone placement on localization accuracy in both near-field and far-field conditions, and having experimentally established the near-field boundary for the apertures considered, we present a demonstration that evaluates how near-field localization accuracy evolves as microphones are progressively deployed during vine robot eversion. Rather than testing pre-assembled static arrays, this demonstration captures the incremental nature of vine robot eversion, where each newly everted microphone extends the array aperture and potentially improves sound source localization. Based on the placement results reported in Section~\ref{subsec:nf_placement}, we selected wall-mounted microphones for this demonstration, as they produced the lowest mean position error ($0.17$~m) and the most consistent estimates across SNR levels under near-field conditions.

\begin{figure}[!t]
    \centering
    \includegraphics[width=0.80\columnwidth]{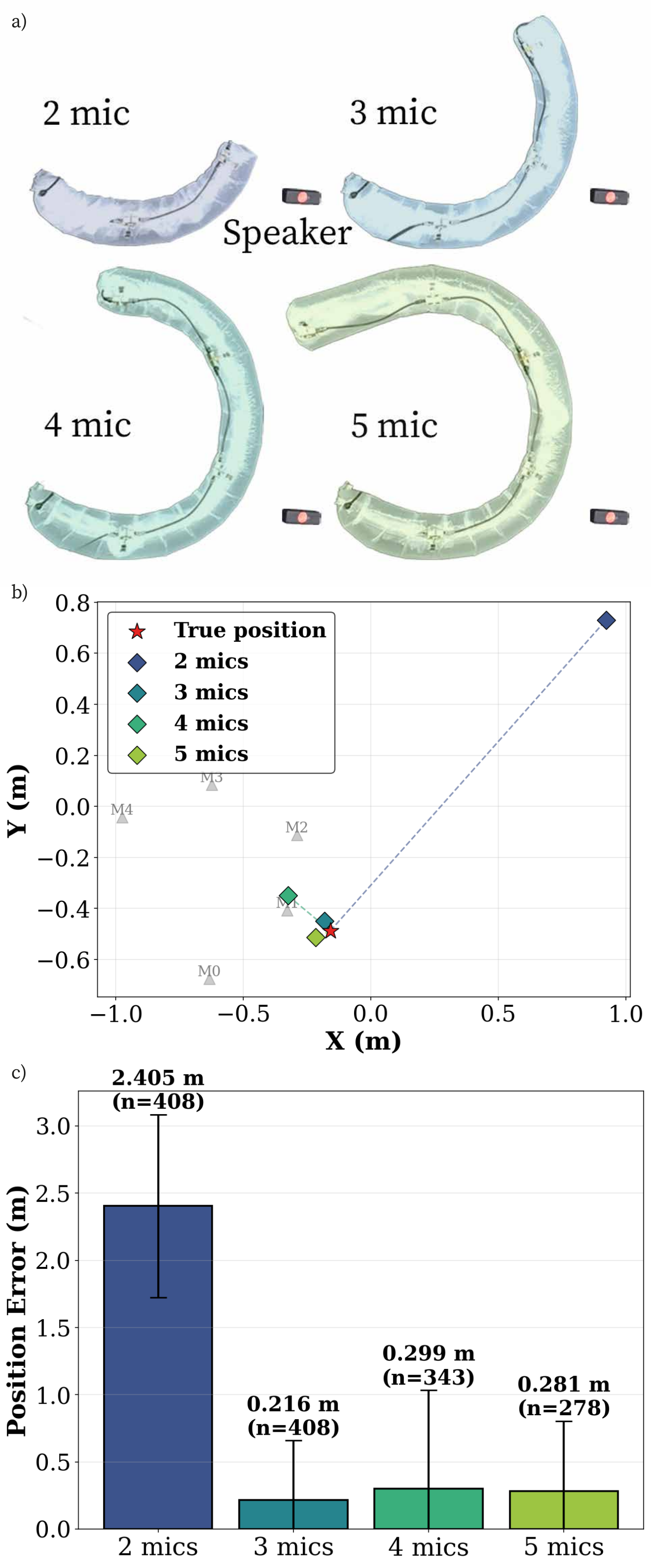}
    
    \caption{System demonstration of a growing vine robot through (a) a time-lapse for each everted microphone, (b) the estimates calculated during each phase as a scatterplot for each incremental configuration, and (c) a bar chart showing the number of samples $n$, mean localization error (m), and standard deviation for each deployment.}
    \label{fig:demonstration}
    \vspace{-1.5em}
\end{figure}

We deployed a semi-circular array with a linearized segment towards the end of the robot, between microphones 4 and 5. Figure~\ref{fig:demonstration}(a) shows a time-lapse of the eversion process, with each microphone color-coded to match the corresponding bars in Figure~\ref{fig:demonstration}(c). We recorded microphone data in a single continuous acquisition throughout the entire eversion sequence. Upon completion, we measured the final position of each microphone and processed the continuous recording, segmenting it according to the timestamps at which each microphone cleared the tail region and became fully everted into its final position. This approach ensures that only frames in which each microphone was deployed and acoustically unobstructed are included in the analysis.

The number of frames available per configuration varies as a consequence of the vine's eversion dynamics: the robot everted slowly during the initial phase, resulting in a longer dwell time at the two-microphone configuration ($n = 408$ frames), followed by three microphones ($n = 408$ frames), four microphones ($n = 343$ frames), and five microphones ($n = 278$ frames). Additionally, we positioned the source $0.5$~m from the center of the array, well within the near-field region established in Section~\ref{subsec:distance}. 

Figure~\ref{fig:demonstration} reports the mean position error for each incremental configuration. The two-microphone array produced the largest error ($2.4$~m), approaching the physical limit imposed by the candidate grid boundary, as a pair of microphones provides insufficient spatial diversity to resolve source range or disambiguate the side from which the sound originates. Adding a third microphone reduced the mean error to $0.216$~m, a reduction of $91.0\%$ relative to the two-microphone baseline
as the additional baseline breaks the left-right ambiguity and introduces wavefront curvature information sufficient for near-field range estimation. The subsequent addition of a fourth and fifth microphone yielded mean errors of $0.299$~m and $0.281$~m, respectively, indicating that the benefit of additional microphones beyond three was marginal and non-monotonic for this particular geometry.

The absence of consistent improvement beyond three microphones, despite the array approaching a more circular geometry as eversion progressed, can be attributed to several concurrent factors. First, the linearization of the array near microphones 4 and 5 partially counteracted the spatial diversity gained from the additional elements. As shown in Section~\ref{subsec:ff_shapes}, collinear microphone segments produce ambiguous TDOA estimates for sources lying near the array axis, reducing the accuracy of the SRP-PHAT map in those angular regions. Second, the near-field placement results in Section~\ref{subsec:nf_placement} demonstrated that wall-mounted microphones achieve their best accuracy when the array geometry maximizes aperture uniformity, a condition that is progressively relaxed as the distal microphones deviate from the circular arc. Third, the decreasing number of available frames at later eversion stages ($n = 343$ and $n = 278$) increases the statistical uncertainty of the mean error estimate, making the differences between the four- and five-microphone configurations less conclusive. Together, these observations suggest that array shape regularity is at least as consequential as microphone count for near-field localization accuracy, and that the geometric trajectory of vine robot eversion should be considered explicitly when planning acoustic sensing deployments in confined spaces.
\section{Conclusions}
\label{sec:conclusions}




In this paper, we presented a novel demonstration of a soft, deformable robot to deploy acoustic sensors. Our approach leverages the length of the robot as it grows to increase the accuracy of sound source localization through the introduction of additional microphones as the robot grows. We also investigated how varying the geometry of the microphone array can aid localization performance. The current work leverages analytical methods, while methods that can incorporate learned or estimated array manifolds, including recent convolutional-based approaches~\cite{diaz2021neural}, offer promising directions for future extension. Our controlled evaluation does not yet capture realistic search-and-rescue conditions such as reverberation, clutter, self-noise, and non-stationary signals, which we leave to future work.

This work establishes a practical demonstration of the feasibility of soft robots to deploy these multi-point acoustic networks. From this set of results, we envision future research into control schemes that actively steer the robot to refine position estimation. Our current approach assumes uniform spacing of microphones along the robot, placed every  30~cm. When building robots of longer lengths, we envision engineering practicalities, such as data and power limits, that complicate such over-instrumentation. Some of these challenges can be overcome by fabricating bespoke sensors that prioritize high-frequency data rates while also minimizing weight-induced limitations to the robot's workspace \cite{mcfarland2025onsteerability}. Overall, this work demonstrates that joint research in the fields of soft robotics and robot audition will enable increased environmental perception, especially in critical field applications such as urban search and rescue.


\bibliographystyle{IEEEtran} 
\bibliography{references}

@article{al2025tip,
  title={Tip-growing robots: Design, theory, application},
  author={Al Harthy, Shamsa and Sadati, SM Hadi and Girerd, C{\'e}dric and Kim, Sukjun and Mondini, Alessio and Wu, Zicong and Saldarriaga, Brandon and Seneci, Carlo A and Mazzolai, Barbara and Morimoto, Tania K and others},
  journal={IEEE Transactions on Robotics},
  year={2025}
}

@inproceedings{bryant2022tactile,
  title={Tactile perception for growing robots via discrete curvature measurements},
  author={Bryant, Micah and Watson, Connor and Morimoto, Tania K},
  booktitle={IEEE/RSJ International Conference on Intelligent Robots and Systems},
  pages={4257--4264},
  year={2022}
}

@inproceedings{der2021roboa,
  title={Ro{B}oa: Construction and Evaluation of a Steerable Vine Robot for Search and Rescue Applications},
  author={der Maur, Pascal Auf and Djambazi, Betim and Haberth{\"u}r, Yves and H{\"o}rmann, Patricia and K{\"u}bler, Alexander and Lustenberger, Michael and Sigrist, Samuel and Vigen, Oda and F{\"o}rster, Julian and Achermann, Florian and others},
  booktitle={IEEE International Conference on Soft Robotics},
  pages={15--20},
  year={2021}
}

@ARTICLE{diaz2021neural,
    title={Robust sound source tracking using {SRP-PHAT} and 3{D} convolutional neural networks},
  author={Diaz-Guerra, David and Miguel, Antonio and Beltran, Jose R},
  journal={IEEE/ACM Transactions on Audio, Speech, and Language Processing},
  volume={29},
  pages={300--311},
  year={2020}
}

@book{dibiase2000srp,
  title={A high-accuracy, low-latency technique for talker localization in reverberant environments using microphone arrays},
  author={DiBiase, Joseph Hector},
  year={2000},
  publisher={Brown University}
}

@INPROCEEDINGS{frost2025rubblesim,
  author={Frost, Constantine and Council, Chad and McGuinness, Margaret and Hanson, Nathaniel},
  booktitle={IEEE International Symposium on Safety Security Rescue Robotics}, 
  title={Rubble{S}im: A Photorealistic Structural Collapse Simulator for Confined Space Mapping}, 
  year={2025},
  pages={215-220}
  }

@INPROCEEDINGS{Fujikawa2019asc,
  author={Fujikawa, Takumi and Yamauchi, Yu and Ambe, Yuichi and Konyo, Masashi and Tadakuma, Kenjiro and Tadokoro, Satoshi},
  booktitle={IEEE International Symposium on Safety, Security, and Rescue Robotics}, 
  title={Development of Practical Air-floating-type Active Scope Camera and User Evaluations for Urban Search and Rescue}, 
  year={2019},
  pages={1-8}
}

@INPROCEEDINGS{Fukuda2014verticalASC,
  author={Fukuda, Junichi and Konyo, Masashi and Takeuchi, Eijiro and Tadokoro, Satoshi},
  booktitle={IEEE/RSJ International Conference on Intelligent Robots and Systems}, 
  title={Remote Vertical Exploration by Active Scope Camera into Collapsed Buildings}, 
  year={2014},
  pages={1882-1888}
  }

@article{grinstein2024ssl,
  title={Steered response power for sound source localization: A tutorial review},
  author={Grinstein, Eric and Tengan, Elisa and {\c{C}}akmak, Bilgesu and Dietzen, Thomas and Nunes, Leonardo and van Waterschoot, Toon and Brookes, Mike and Naylor, Patrick A},
  journal={EURASIP Journal on Audio, Speech, and Music Processing},
  volume={2024},
  number={1},
  pages={59},
  year={2024},
  publisher={Springer}
}

@inproceedings{gruebele2021distributed,
  title={Distributed sensor networks deployed using soft growing robots},
  author={Gruebele, Alexander M and Zerbe, Andrew C and Coad, Margaret M and Okamura, Allison M and Cutkosky, Mark R},
  booktitle={IEEE International Conference on Soft Robotics},
  pages={66--73},
  year={2021}
}

@article{hegde2023sensing,
    title={Sensing in soft robotics},
  author={Hegde, Chidanand and Su, Jiangtao and Tan, Joel Ming Rui and He, Ke and Chen, Xiaodong and Magdassi, Shlomo},
  journal={ACS Nano},
  volume={17},
  number={16},
  pages={15277--15307},
  year={2023},
  publisher={ACS Publications}
}

@INPROCEEDINGS{keyrouz2006binaural,
  title={An enhanced binaural 3{D} sound localization algorithm},
  author={Keyrouz, Fakheredine and Diepold, Klaus},
  booktitle={IEEE International Symposium on Signal Processing and Information Technology},
  pages={662--665},
  year={2006}
}

@ARTICLE{knapp1976gcc,
  title={The generalized correlation method for estimation of time delay},
  author={Knapp, Charles and Carter, Glifford},
  journal={IEEE Transactions on Acoustics, Speech, and Signal Processing},
  volume={24},
  number={4},
  pages={320--327},
  year={1976}
}

@article{laudenslager2026evaluating,
  title={Evaluating Accuracy of Vine Robot Shape Sensing with Distributed Inertial Measurement Units},
  author={Laudenslager, Alexis E and Valdivia, Antonio Alvarez and Hanson, Nathaniel and McGuinness, Margaret},
  journal={arXiv preprint arXiv:2602.24202},
  year={2026}
}

@article{Lee2017SoftRobotReview,
  title={Soft robot review},
  author={Lee, Chiwon and Kim, Myungjoon and Kim, Yoon Jae and Hong, Nhayoung and Ryu, Seungwan and Kim, H Jin and Kim, Sungwan},
  journal={International Journal of Control, Automation and Systems},
  volume={15},
  number={1},
  pages={3--15},
  year={2017},
  publisher={Springer}
}

@inproceedings{mcfarland2024field,
  title={Field insights for portable vine robots in urban search and rescue},
  author={McFarland, Ciera and Dhawan, Ankush and Kumari, Riya and Council, Chad and Coad, Margaret and Hanson, Nathaniel},
  booktitle={IEEE International Symposium on Safety Security Rescue Robotics},
  pages={190--197},
  year={2024}
}

@article{mcfarland2025onsteerability,
  title={On Steerability Factors for Growing Vine Robots},
  author={McFarland, Ciera and Alvarez Valdivia, Antonio and Taher, Sarah and Hanson, Nathaniel and McGuinness, Margaret},
  journal={arXiv preprint arXiv:2510.22504},
  year={2025}
}

@inproceedings{mitchell2023soft,
  title={Soft air pocket force sensors for large scale flexible robots},
  author={Mitchell, Michael R and McFarland, Ciera and Coad, Margaret M},
  booktitle={IEEE International Conference on Soft Robotics},
  pages={1--8},
  year={2023}
}

@book{murphy2017disaster,
  title={Disaster Robotics},
  author={Murphy, Robin R},
  year={2017},
  publisher={MIT Press}
}

@INPROCEEDINGS{okuno2015audition,
  title={Robot audition: Its rise and perspectives},
  author={Okuno, Hiroshi G and Nakadai, Kazuhiro},
  booktitle={IEEE International Conference on Acoustics, Speech and Signal Processing},
  pages={5610--5614},
  year={2015}
}

@article{qu2024sensing,
  title={Advanced flexible sensing technologies for soft robots},
  author={Qu, Juntian and Cui, Guangming and Li, Zhenkun and Fang, Shutong and Zhang, Xianrui and Liu, Ang and Han, Mingyue and Liu, Houde and Wang, Xueqian and Wang, Xiaohao},
  journal={Advanced Functional Materials},
  volume={34},
  number={29},
  pages={2401311},
  year={2024},
  publisher={Wiley Online Library}
}

@article{roy1989ssl_esprit,
  title={{ESPRIT}-estimation of signal parameters via rotational invariance techniques},
  author={Roy, Richard and Kailath, Thomas},
  journal={IEEE Transactions on Acoustics, Speech, and Signal Processing},
  volume={37},
  number={7},
  pages={984--995},
  year={1989}
}

@inproceedings{sasaki2018online,
  title={Online spatial sound perception using microphone array on mobile robot},
  author={Sasaki, Yoko and Tanabe, Ryo and Takernura, Hiroshi},
  booktitle={IEEE/RSJ International Conference on Intelligent Robots and Systems},
  pages={2478--2484},
  year={2018}
}

@book{schmidt1982ssl_music,
  title={A signal subspace approach to multiple emitter location and spectral estimation},
  author={Schmidt, Ralph Otto},
  year={1982},
  publisher={Stanford University}
}

@article{thuruthel2019shapesensing,
  title={Soft robot perception using embedded soft sensors and recurrent neural networks},
  author={Thuruthel, Thomas George and Shih, Benjamin and Laschi, Cecilia and Tolley, Michael Thomas},
  journal={Science Robotics},
  volume={4},
  number={26},
  pages={eaav1488},
  year={2019},
  publisher={American Association for the Advancement of Science}
}

@inproceedings{valin2003robust,
  title={Robust sound source localization using a microphone array on a mobile robot},
  author={Valin, J-M and Michaud, Fran{\c{c}}ois and Rouat, Jean and L{\'e}tourneau, Dominic},
  booktitle={IEEE/RSJ International Conference on Intelligent Robots and Systems (Cat. No. 03CH37453)},
  volume={2},
  pages={1228--1233},
  year={2003}
}

@inproceedings{yang2024kidnappable,
  title={The un-kidnappable robot: Acoustic localization of sneaking people},
  author={Yang, Mengyu and Grady, Patrick and Brahmbhatt, Samarth and Vasudevan, Arun Balajee and Kemp, Charles C and Hays, James},
  booktitle={IEEE International Conference on Robotics and Automation},
  pages={985--992},
  year={2024}
}

@ARTICLE{adavanne2018,
  author={Adavanne, Sharath and Politis, Archontis and Nikunen, Joonas and Virtanen, Tuomas},
  journal={IEEE Journal of Selected Topics in Signal Processing}, 
  title={Sound Event Localization and Detection of Overlapping Sources Using Convolutional Recurrent Neural Networks}, 
  year={2018},
  volume={13},
  number={1},
  pages={34-48},
  doi={10.1109/JSTSP.2018.2885636}}





\end{document}